# Intuitive sequence matching algorithm applied to a sip-and-puff control interface for robotic assistive devices

Frédéric Schweitzer[1,2], Alexandre Campeau-Lecours[1,2]

[1]*Department of Mechanical Engineering, Université Laval, Quebec City, Canada,* [2]*Centre for Interdisciplinary Research in Rehabilitation and Social Integration (CIRRIS), CIUSSSCN, Quebec City, Canada*

## ABSTRACT

This paper presents the development and preliminary validation of a control interface based on a sequence matching algorithm. An important challenge in the field of assistive technology is for users to control high dimensionality devices (e.g., assistive robot with several degrees of freedom, or computer) with low dimensionality control interfaces (e.g., a few switches). Sequence matching consists in the recognition of a pattern obtained from a sensor's signal compared to a predefined pattern library. The objective is to allow the user to input several different commands with a low dimensionality interface (e.g., Morse code allowing inputting several letters with a single switch). In this paper, the algorithm is used in the context of the control of an assistive robotic arm and has been adapted to a sip-and-puff interface where short and long bursts can be detected. Compared to a classic sip-and-puff interface, a preliminary validation with 8 healthy subjects has shown that sequence matching makes the control faster, easier and more comfortable. This paper is a proof of concept that leads the way towards a more advanced algorithm and the usage of more versatile sensors such as inertial measurement units (IMU).

## INTRODUCTION

Interactions between human and technology is well documented and known as "human-robot interactions" in industrial robotics. When it comes to assistive technology (AT), the importance to develop around this aspect of engineering is high and the human part becomes predominant. On the one hand, a device like an assistive robotic arm can help a person gain more autonomy in their daily life [1]. However, the control interface (the link between the human operator and the AT) is often portrayed as the system's Achilles heel preventing users to efficiently control their AT [2, 3]. This is amplified by the fact that each user has his/her own abilities and may require specific control interface for his/her condition. For instance, controlling a six-degree-of-freedom robotic arm or a computer with a low dimensional interface is not always an easy task. Assistive robotic arm users need to control many degrees of freedom and components (translations, rotations, fingers, etc.) with only a two-axis joystick and switches[4]. Other users rely solely on a sip-and-puff control interface, which makes the control even more difficult since they only use two inputs (inhale and exhale) to control the many degrees of freedom. These users must thus navigate through mode selection to activate different control modes, which can be a time-consuming, demanding and sometimes impossible process.

In recent years, alternative control interfaces have emerged for ATs. Among the alternatives, assuming the user can speak clearly enough, voice control interfaces represent a valuable option. Commercial voice assistants can help people living with disabilities in various domestic tasks such as making phone calls, playing music, or handling lights [2]. Although these assistants require a constant Internet connection, it's also possible to develop offline interfaces [5]. Some people prefer control with IMUs. Users who can move their head and/or lower limbs easily can wear IMUs and control a robot [6] or a computer mouse [7]. Electromyography (EMG) interfaces may also be an option [8] in specific cases. Finally, sip-and-puff or tongue control devices [9] are widely used but are such low dimensional interface that they represent a big challenge for the user. These devices can be more efficiently interfaced with an AT using intelligent algorithms. To that effect, automatic mode switching techniques are in development [3], but may have their limitations for now. In addition, these algorithms are well suited for robotic arms, but the interface should also be adaptable to different ATs.

Some of these emerging control interfaces are somewhat cumbersome in practice and may require many calibrations by the user. This may explain why standard interfaces such as joysticks and sip-and-puff devices are predominant to this day in practice.

## OBJECTIVES

The objective of this project is to develop and evaluate an intuitive control interface based on currently used devices (e.g., switches and joysticks) and to combine them with intelligent algorithms to enhance the user's control interface and help people living with disabilities to control their assistive devices. This paper presents a sequence matching algorithm as a proof of concept, and shows that using low dimensionality sensors with intelligent signal recognition algorithms can improve the control of ATs.

## SEQUENCE MATCHING ALGORITHM

This section presents the proposed sequence matching algorithm. The general idea behind the development of this algorithm is to detect a sequence of actions executed by the user and to map those sequences to the controls of assistive devices (e.g., mode selection of a robotic arm, keyboard shortcuts on a computer, home automation). For a sequence to be matched, it must correspond to a preset user-defined sequence (UDS). The sequence to match can vary from simple units (short or long presses) to more complex actions (e.g., comparing two continuous analog signals at 1 kHz). In the specific context of this paper, the algorithm was applied to a sip-and-puff sensor with four possible actions: short sip (inhaling), long sip (inhaling), short puff (exhaling), long puff (exhaling). The actions are defined in the software respectively as the following signals: 1, 2, -1 and -2. In this paper, the signals are used to control an assistive robotic arm with several degrees of freedom. The available modes are: 1) forward-backward translations, 2) left-right translations, 3) up-down translations, 4) *x*-axis rotation, 5) *y*-axis rotation, 6) and *z*-axis rotation, 7) open-close fingers, 8) save a point and 9) go to a saved point.

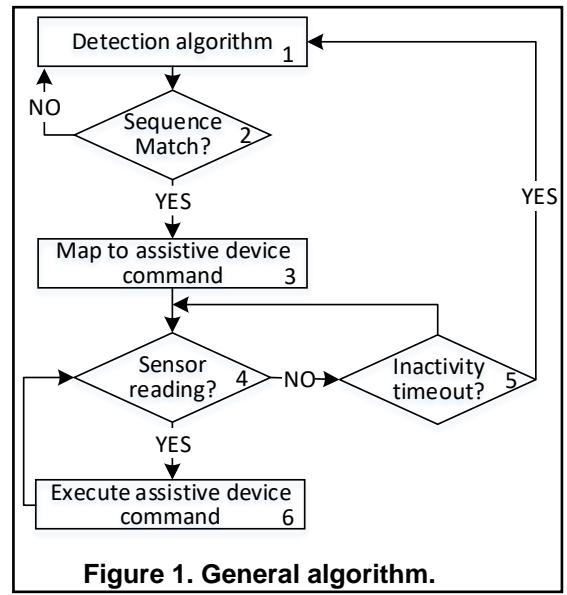

**Figure 1. General algorithm.**

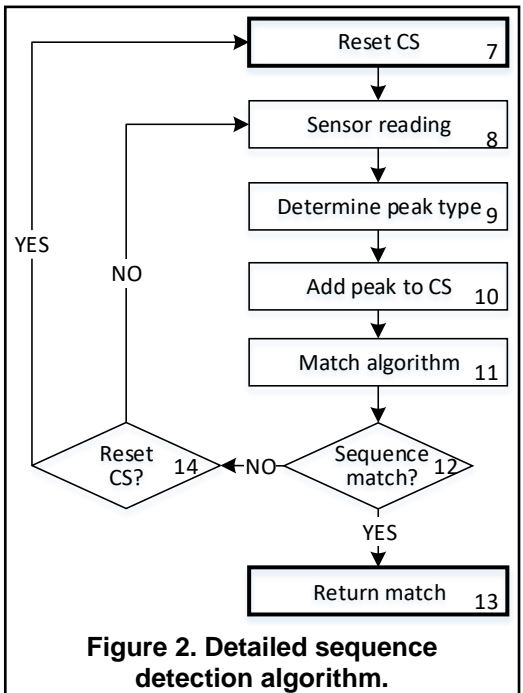

**Figure 2. Detailed sequence detection algorithm.**

Figure 1 is a visual representation of the general algorithm. In the following description, numbers in parentheses refer to a corresponding block in Figures 1, 2 and 3. The program starts in detection mode (1) and remains in that mode until the current sequence (CS) matches one of the user-defined sequences. When this happens, the robot is ready to receive a command (i.e., moving according to the selected mode) (3). For instance, if the mode forward-backward has been selected, the robot will move forward with an exhalation and backwards with an inhalation (6). The command mapping can also be user-defined, meaning that the user can tell the system which sequence corresponds to which mode. After a moment of inactivity (no inhale or exhale) (5), the algorithm returns to detection mode.

The mode detection algorithm is detailed in Figure 2. The algorithm acquires sensor data (8) to detect peaks and determine their nature (9). A peak can be short or long and up (exhale) or down (inhale). It is detected with thresholds on the 0-5 V signal. The time difference between the moment the signal goes up and the moment it comes down (or the other way around) determines if it is long or short. Detected peaks are stacked in a list (either as -2, -1, 2, 2) (10) until a UDS is matched (14) or the list is reset.

Figure 3 shows the detailed match algorithm which task is to determine what to do with the CS. For instance, given the UDSs: $S_1 = \langle 1, 2, -1 \rangle$, $S_2 = \langle 1, 2 \rangle$, $S_3 = \langle 2, 1 \rangle$, if the CS (15) inputted by the user is $S_0 = \langle 2 \rangle$, there is no confusion (16-17-19-22), but the program waits for a 1 to match $S_3$ (16-17-18-21). Otherwise, the CS is reset if the user takes too much time to complete any UDS (16-20-23). If $S_0 = \langle 1, 2 \rangle$ instead, there is confusion between $S_1$ and $S_2$ (16-17-18-22). The program has to wait to see if the user sends a new peak. If so, it shall correspond to -1 to match $S_1$ (16-17-18-21) otherwise the CS is reset (16-17-19-23) as it was not a viable command. If no new peaks are sent after $\langle 1, 2 \rangle$, $S_2$ is matched (16-20-21). This method ensures that the compatible sequences are automatically matched and the user does not lose any time waiting for his sequence to be matched. In addition, the waiting timers (5, 16) may be adjusted to the user to increase or decrease the waiting time. The whole algorithm was adapted to a sip-and-puff sensor, but the sensor part (8) can be replaced by any sensor that best suit the user's abilities.

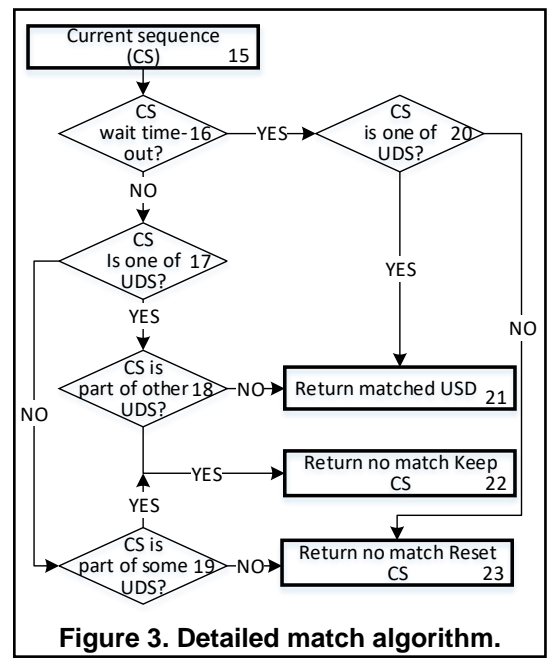

**Figure 3. Detailed match algorithm.**

## EXPERIMENTATION

As noted in the previous section, the algorithm has been adapted for a sip-and-puff interface for the control of an assistive robotic arm. In the experiments, a JACO assistive robotic arm [10] produced by Kinova Robotics was used. In order to assess the performance of the proposed control interface, three tasks have been performed. Each task was performed with both a classic sip-and-puff (named Basic Sip-and-Puff (BSP)) control for assistive robots and with proposed control algorithm (named Advanced Sip-and-Puff (ASP)). The BSP is the more common and commercialized sip-and-puff interface. It allows the user to view all modes auto-scrolling on the screen. When the desired mode is highlighted, the user exhales once to enter the mode and then inhales or exhales to go back and forth in this mode. The ASP shows a graphical interface where a sequence is displayed beside each corresponding mode (e.g., sip-sip-sip = forward-backward).

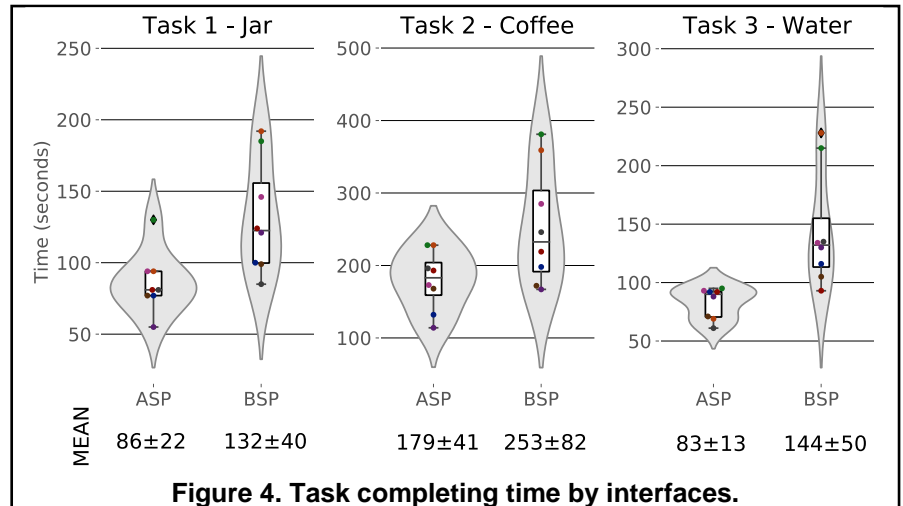

**Figure 4. Task completing time by interfaces.**

Eight healthy participants (6 males and 2 females) aged between 23 and 33 years old took part in the experiments. Half the participants began with the ASP and the other half with the BSP.

The first task consisted in grabbing a jar on a shelf and to bring it and drop it on a table. The second task was to grab a plastic block holding a spoon from a shelf, take a spoonful of coffee and put it in a mug on a table. The third task consisted in taking a bottle of water on a shelf and pour water in a glass on a table. Those tasks are inspired by the work of Beaudoin et al. [11]. All tasks are within the reach of the robotic arm.

Before the experiment begins, the participants were allowed a ten-minute familiarization time with each control mode. The participants were timed for each task (full completion and moving time). Figure 4 shows all the task completion times of the participants. After the tests, the participants were asked to answer a questionnaire, which contained questions adapted from the QUEAD of Schmidtler et al. [12], and leave their comments.

## RESULTS AND DISCUSSION

Figure 4, shows box plots, estimated probability densities and means, plus or minus the standard deviation (SD) of the times for the full completion of the three tasks by interface. Each point of different color on the plots corresponds to a distinct participant. The completion times with the ASP is significantly lower than the completion times with the BSP according to one-tailed, non-parametric Wilcoxon signed-rank tests (for paired data) for each task ($p$-value = $5.86 \times 10^{-3}$ > 0.05, for each task). Indeed, for tasks 1 to 3, the ASP completion times are respectively 35%, 30% and 42% lower than the BSP completion times.

**Table 1: Post experiment questionnaire**

| | Completely disagree | Mostly disagree | Neutral | Mostly agree | Completely agree |
|---|---|---|---|---|---|
| **Perceived usefulness** | | | | | |
| The ASP interface is useful | | | | 1 | 7 |
| The ASP interface enhances my performance | | | | 1 | 7 |
| I could efficiently complete the tasks using the ASP interface | | | | 4 | 4 |
| I could efficiently complete the tasks using the BSP interface | | 1 | 3 | 2 | 2 |
| **Perceived ease of use** | | | | | |
| The ASP interface is easy to use | | | | 5 | 3 |
| The ASP interface is easier to use than the BSP interface | | | 2 | 1 | 5 |
| The ASP interface feels cumbersome to use | 3 | 3 | 1 | 1 | |
| The BSP interface feels cumbersome to use | | 1 | | 6 | 1 |
| I didn't need concentration to use the ASP interface | | 3 | 2 | 3 | |
| I didn't need concentration to use the BSP interface | 1 | 2 | | 5 | |
| Using the ASP interface was intuitive | | 1 | 1 | 3 | 3 |
| **Emotions** | | | | | |
| I prefer using the ASP interface rather than the BSP interface | | | | | 8 |
| **Attitude** | | | | | |
| In a long-term perspective, I think it would be faster to use the ASP interface | | | | | 8 |
| In a long-term perspective, I think it would be difficult to learn multiple command of the ASP interface | 6 | 2 | | | |

Since the tasks are the same for both interfaces, the moving time (the time the robot actually moves) should also be the same. This is confirmed by a two-tailed Wilcoxon test on the moving times (again for paired data). The test fails to reject the null hypothesis. Therefore there is no evidence to assume that the moving times are different between both interfaces ([$p$-value = 0.433] > 0.05, on ASP – BSP difference: mean = 0.54, SD = 3.80 seconds). The extra time from ASP to BSP is therefore wasted time during which the participants wait for the interface to keep up. This is consistent with the fact that the variances of estimated probability densities are higher for BSP than ASP on Figure 4. This strong variability shows that the participants made more mistakes (i.e., missing a mode during the auto-scroll) using BSP, and that those mistakes cost time.

The results of the questionnaire are presented in Table 1. In general, participants found the ASP more useful and easier to use compared to the BSP and they preferred using the ASP. On the other hand, the participants thought using ASP required more concentration than using BSP. However, they also mentioned that, with an adaptation and learning period, the ASP would become much more comfortable to use. This leads to conclude that the difference between both interfaces is the control mode selection, which is less time-consuming with the ASP.

**CONCLUSION**

Controlling assistive technologies of high dimensionality with a low dimension interface is time- and energy-consuming for many users. This paper has presented a sequence matching algorithm that was applied to the control of an assistive robotic arm with a sip-and-puff interface. The experiments conducted show that the algorithm made the control faster and the interface more comfortable. This preliminary validation with healthy subjects is encouraging. The next steps will consist in validating the algorithm with actual assistive robot users along as to develop a new version of the algorithm that can work with continuous signals obtained from different sensors such as inertial measurement units in order to detect custom movements.

**ACKNOWLEDGEMENT**

This work is supported by Dr. Campeau-Lecours's startup funds at CIRRIS and Université Laval.